\documentclass[10pt,twocolumn,letterpaper]{article}

\usepackage{ijcb}
\usepackage{times}
\usepackage{epsfig}
\usepackage{graphicx}
\usepackage{amsmath}
\usepackage{amssymb}
\usepackage{color}
\usepackage[colorlinks=true, citecolor={blue}, bookmarks=false, breaklinks]{hyperref}
\usepackage{multirow}
\usepackage{adjustbox}
\usepackage{url}

\usepackage{breakurl}
\usepackage{algorithm}
\usepackage[noend]{algpseudocode}
\usepackage{bm}
\newcommand{\R}{\mathbb{R}}

\ijcbfinalcopy % *** Uncomment this line for the final submission

\ifijcbfinal\fi
\begin{document}
	
%%%%%%%%% TITLE
\title{Generating Master Faces for Use in Performing Wolf Attacks on \\ Face Recognition Systems}

\author{Huy H. Nguyen$^{1}$, Junichi Yamagishi$^{1,2,4}$, Isao Echizen$^{1,2,3}$, and S\'ebastien Marcel$^{5}$\\
$^{1}$The Graduate University for Advanced Studies, SOKENDAI, Kanagawa, Japan\\
	$^{2}$National Institute of Informatics, Tokyo, Japan; $^{3}$The University of Tokyo, Japan\\
	$^{4}$The University of Edinburgh, Edinburgh, UK; $^{5}$Idiap Research Institute, Martigny, Switzerland\\
	\small{Email: \{nhhuy, jyamagis, iechizen\}@nii.ac.jp, marcel@idiap.ch}
}

\maketitle
\thispagestyle{empty}
	
%%%%%%%%% ABSTRACT
\begin{abstract}
	Due to its convenience, biometric authentication, especial face authentication, has become increasingly mainstream and thus is now a prime target for attackers. Presentation attacks and face morphing are typical types of attack. Previous research has shown that finger-vein- and fingerprint-based authentication methods are susceptible to wolf attacks, in which a wolf sample matches many enrolled user templates. In this work, we demonstrated that wolf (generic) faces, which we call ``master faces,'' can also compromise face recognition systems and that the master face concept can be generalized in some cases. Motivated by recent similar work in the fingerprint domain, we generated high-quality master faces by using the state-of-the-art face generator StyleGAN in a process called latent variable evolution. Experiments demonstrated that even attackers with limited resources using only pre-trained models available on the Internet can initiate master face attacks. The results, in addition to demonstrating performance from the attacker's point of view, can also be used to clarify and improve the performance of face recognition systems and harden face authentication systems.
\end{abstract}

%%%%%%%%% BODY TEXT
\section{Introduction}
\label{sec:intro}

Recent advances in the development of biometric authentication, especially in its ease of use, have enabled face authentication (which uses face recognition) to be implemented in many portable and handheld devices, from laptop PCs to smartphones. Digital wallets, which are also called ``e-wallets'' and are popular in many countries, also utilize face authentication from the user's smartphone to process payments. As a result, face authentication systems have become a prime target for attackers. Even before this trend, interest in creating a face that matches multiple faces led researchers to come up with the idea of face morphing~\cite{scherhag2019face}, which is a special case of image morphing~\cite{wolberg1998image}. Given two or more faces from different identities, a system creates a blended face that can match all component identities when using face recognition (FR) systems and possibly fool a human observer. Morphing attacks often target automated border control systems, possibly by criminals to avoid being detected~\cite{scherhag2019face}.

\begin{figure}[t]
	\centering
	\includegraphics[width=\columnwidth]{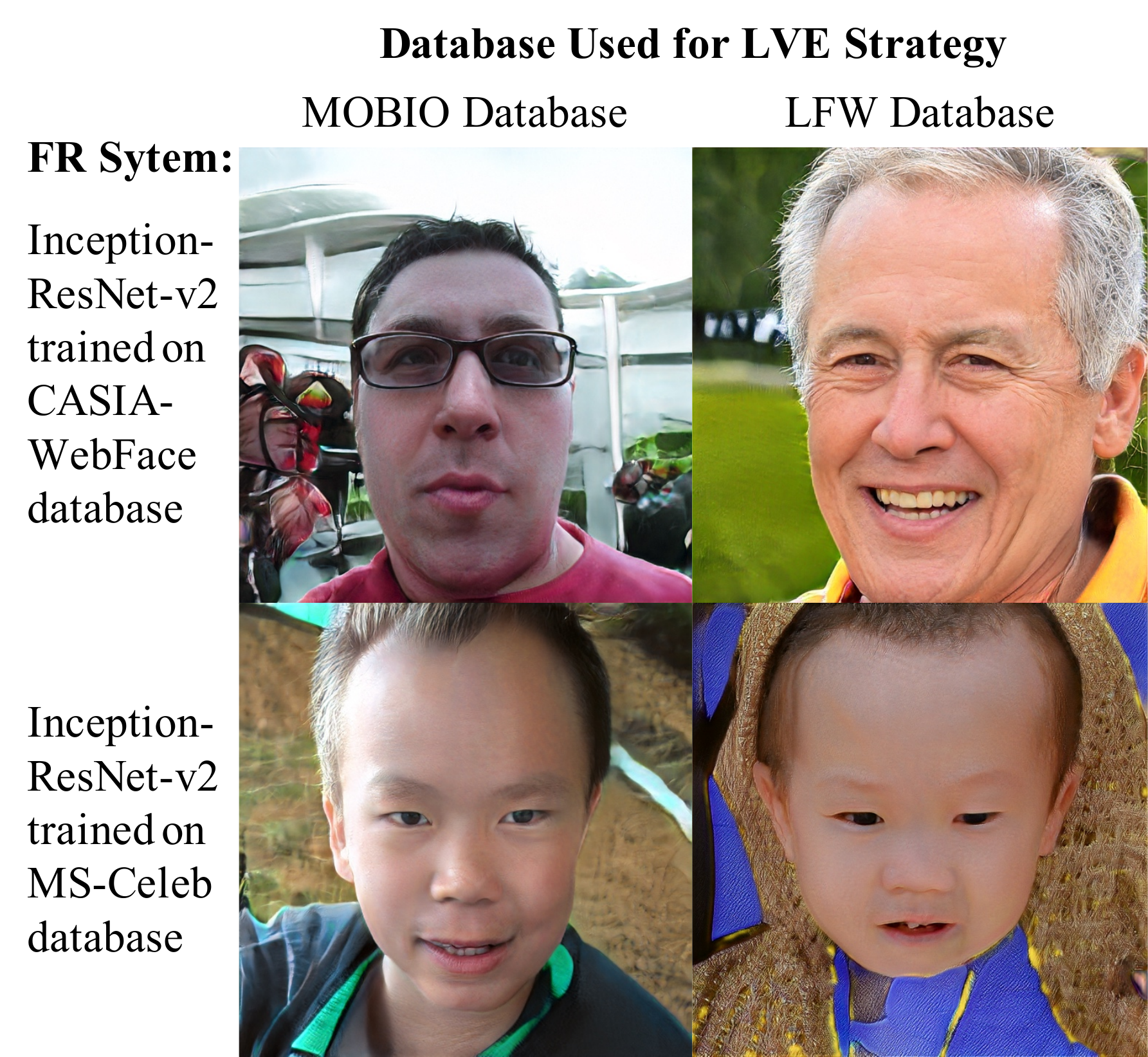}
	\caption{Example master faces generated with our method. Images in first and second column were generated using the MOBIO~\cite{mccool2012bi} and Labeled Faces in the Wild~\cite{LFWTechUpdate} databases, respectively. Training for images in first row used Inception-ResNet-v2~\cite{szegedy2017inception} based FR system trained on CASIA-WebFace database~\cite{yi2014learning} while that for ones in second row used same FR system trained on MS-Celeb database~\cite{guo2016ms}.}
	\label{fig:masterfaces}
\end{figure}

Another kind of attack is called a ``wolf attack,'' which targets biometric authentication systems by finding or crafting a generic sample, a ``wolf sample,'' that has high similarity to many of the enrolled templates~\cite{une2007wolf}. The advantage of a wolf attack is that it does not require knowledge about the enrolled subjects. Wolf attacks are commonly aimed at finger-vein- and fingerprint-based recognition systems~\cite{une2007wolf,bontrager2018deepmasterprints}. Bontrager et al. introduced a generative adversarial network (GAN)~\cite{goodfellow2014generative} based method for generating realistic fingerprint images (``MasterPrints'') using the latent variable evolution (LVE) strategy for use in attacking systems using partial fingerprint images~\cite{bontrager2018deepmasterprints}. Although a defense method was subsequently proposed~\cite{roy2019masterprint}, this kind of attack has much room for improvement.

In this work, we demonstrated that wolf attacks can also compromise FR systems. An LVE algorithm running on a pre-trained high-quality face generator StyleGAN~\cite{karras2019style} was able to match selected \textbf{master face} (wolf) samples with multiple user templates (in-domain and out-of-domain) for both known and unknown FR systems. A \textbf{known} FR system is the one used on the running LVE algorithm while an \textbf{unknown} FR system can be understood as one with the same architecture but trained on a different database or one with a completely different architecture. We modified the LVE algorithm~\cite{bontrager2018deepmasterprints} by changing the way it calculates the scores of the generated faces. The master faces can be easily and quickly generated by simply using pre-trained models available on the Internet. Examples are shown in Figure~\ref{fig:masterfaces}.

We used a combination of a StyleGAN model pre-trained on the Flickr-Faces-HQ (FFHQ) database~\cite{karras2019style} and an Inception-ResNet-v2~\cite{szegedy2017inception} based FR system pre-trained by de Freitas Pereira et al.~\cite{de2018heterogeneous} on the CASIA-WebFace database~\cite{yi2014learning} (available in Bob toolbox~\cite{bob2017}) to generate the master faces. With this combination and less than 24 hours of training on a conventional personal computer (PC) without a graphics processing unit (GPU), we generated master faces that achieved false acceptance rates (FARs) between 6 and 35\% depending on the test database and targeted FR system. These high rates raise a major concern about the ability of FR systems to deal with a master face attack, which can be launched by someone without any special training. In addition to considering the attacker's point of view, we also consider countermeasures to mitigate this threat.

The rest of the paper is organized as follows. First, we present general information about facial image generation, FR systems, wolf attacks, and the LVE algorithm in section~\ref{sec:related}. We then introduce our proposed method in section~\ref{sec:proposed} and describe our experimental design, present the results, and discuss them in section~\ref{sec:eval}. Next, we discuss possible ways in the literature to defend against master face attacks in section~\ref{sec:defense}. Finally, we summarize the key points and mention future work in section~\ref{sec:sum}.  

\section{Related Work}
\label{sec:related}
\subsection{Facial Image Generation}

Facial image generation has recently been attracting the attention of the research community, especially since the introduction of variational autoencoders (VAEs)~\cite{kingma2014auto} and GANs~\cite{goodfellow2014generative}. Initially, only facial images with low resolution and sizes were generated. In addition, VAEs suffer a trade-off between disentangled representations and reconstruction errors while GANs are difficult to train. To solve the later problem, Arjovsky et al. proposed the Wasserstein GAN (WGAN), which improves the stability of learning and eliminates the mode collapse problem~\cite{arjovsky2017wasserstein}. Subsequent work led to an improved WGAN called WGAN-GP in which the weight clipping is replaced with a gradient-based penalty function~\cite{gulrajani2017improved}. WGAN~\cite{arjovsky2017wasserstein,gulrajani2017improved} was used in the work of Bontrager et al. to generate master prints for use in attacking partial fingerprint authentication systems~\cite{bontrager2018deepmasterprints}.

Despite these improvements, GANs still suffer the problem of generating high-resolution images. Karras et al. proposed a training methodology in which both the generator and discriminator are progressively trained~\cite{karras2018progressive}. It starts with a low-resolution image model and then repeatedly adds new layers to the model to incorporate fine details. Using this idea of progressive training and borrowing the idea of image generating from the style transfer field, Karras et al. introduced a novel face generator network called StyleGAN~\cite{karras2019style}. This network has the ability to automatically learn and separate high-level attributes (such as pose and identity) and stochastic variation in the generated images (such as freckles and hair). Unlike traditional GANs, StyleGAN includes two components: a mapping network that maps the input latent vector to intermediate style vectors and feeds them into the synthesis network. With these two components, StyleGAN handles disentanglement well and supports style mixing. Subsequent work focused on analyzing and improving the quality of StyleGAN generated imags~\cite{karras2019analyzing}. In this work, we used StyleGAN~\cite{karras2019style} for facial image generation.

\subsection{Face Recognition Systems}

Recent advances in convolutional neural networks (CNNs) and the releases of large annotated databases substantially improved the performances of FR systems, enabling them to be applied to not only homogeneous but also heterogeneous domains~\cite{de2018heterogeneous}. Two examples of such large databases are the CASIA-WebFace database~\cite{yi2014learning} and the MS-Celeb database~\cite{guo2016ms}, which are commonly used to create training data for state-of-the-art FR systems. Smaller well-known databases that had been previously released, like the MOBIO database~\cite{mccool2012bi} and the Labeled Faces in the Wild (LFW) database~\cite{LFWTechUpdate}, are usually used for validation.
Reusing an architecture that performed well in the ImageNet Large Scale Visual Recognition Challenge (ILSVRC)~\cite{ILSVRC15} as a feature extractor for CNN-based FR systems, rather than designing a new architecture from scratch, is a commonly used approach. The two most commonly used architectures are the VGG (Visual Geometry Group) network~\cite{simonyan2014very} and the Inception network~\cite{szegedy2017inception}.

Parkhi et al. adopted the VGG-16 network~\cite{simonyan2014very} for an FR task (``VGG-Face'') and trained it on a custom-built large-scale database~\cite{parkhi2015deep}. Wu et al. introduced a light CNN that uses max-feature-map activation and has ten times fewer parameters than the VGG-Face network~\cite{wu2018light}. The Inception architecture~\cite{szegedy2017inception} was used by Schroff et al. to build the FaceNet model, which maps facial images to a compact Euclidean space embedding~\cite{schroff2015facenet}. Therefore, it can be used for face verification, recognition, and clustering. The closet open-source implementation of FaceNet was done by David Sandberg~\cite{sandberg2017facenet} using the Inception-ResNet v1 and v2 architectures~\cite{szegedy2017inception}.
Additionally, de Freitas Pereira et al. used the Inception-ResNet v2 architecture in their heterogeneous FR work~\cite{de2018heterogeneous}. Experiments demonstrated that Inception-based methods perform better than VGG-based ones. Using another approach, Tran et al. proposed a disentangled representation learning GAN method (``DR-GAN'') that can deal with face variations, especially in pose~\cite{tran2017disentangled}.
In our experiments, we used three FR systems: (1) Inception-ResNet v1 based FaceNet by David Sandberg~\cite{sandberg2017facenet}, (2) Inception-ResNet v2 network by de Freitas Pereira et al.~\cite{de2018heterogeneous}, and (3) DR-GAN by Tran. et al.~\cite{tran2017disentangled}.

\subsection{Wolf Attacks}

For biometric authentication systems, Une et al.~\cite{une2007wolf} defined a ``wolf sample'' as an input sample that can be falsely accepted as a match with multiple user templates (enrolled subjects). They also defined a measurement called wolf attack probability (WAP), which is the maximum probability of a successful attack with one wolf sample. Wolf attacks and defenses against them are the subject of much research in the finger-vein and fingerprint recognition fields. Wolf attacks has been evolved from generating forged minutiae~\cite{ratha2001enhancing} to generating real partial fingerprint images~\cite{bontrager2018deepmasterprints}. In the latest work~\cite{bontrager2018deepmasterprints}, Bontrager et al. used a latent variable evolution algorithm to maximize the WAPs of partial fingerprint images generated by a GAN. This type of attack is applicable to systems using small-size sensors with limited resolution. In contrast, in this work, we focused on generating high-quality high-resolution facial images for use in attacking FR systems using full-face input.

\subsection{Latent Variable Evolution Algorithm}
Inspired by biological evolution, evolution algorithms have long been used in artificial intelligence applications without any assumption about the underlying fitness landscape. One such algorithm is the evolution strategies (ES) algorithm, which can be used for complex, multimodal, and non-differentiable functions. Proposed by Hansen and Ostermeier, the Covariance Matrix Adaptation Evolution Strategy (CMA-ES) is a powerful ES algorithm designed for non-linear and non-convex functions~\cite{hansen2001completely}.
Bontrager et al. demonstrated that interactive evolutionary computation can be used in combination with a GAN~\cite{bontrager2018deep}. After the GAN is trained, a latent vector used as input to the GAN can be put under evolutionary control, resulting in the generation of high-quality samples. Bontrager et al. proposed combining this method with the CMA-ES algorithm~\cite{hansen2001completely} to generate partial fingerprints. The resulting LVE algorithm maximizes the WAP of the generated partial fingerprint images against a fingerprint authentication system~\cite{bontrager2018deepmasterprints}.
Following this success in the fingerprint domain, we modified the scoring method used for the LVE algorithm and applied the resulting algorithm to the facial domain, which is trickier since human vision is more sensitive to faces than fingerprints. With the help of the StyleGAN high-quality face generator~\cite{karras2019style} and the powerful CMA-ES algorithm~\cite{hansen2001completely}, our proposed method can generate high-resolution master faces that are both natural looking and have a high WAP.

\section{Proposed Method}
\label{sec:proposed}

\subsection{Overview}
\begin{figure}[th!]
	\centering
	\includegraphics[width=\columnwidth]{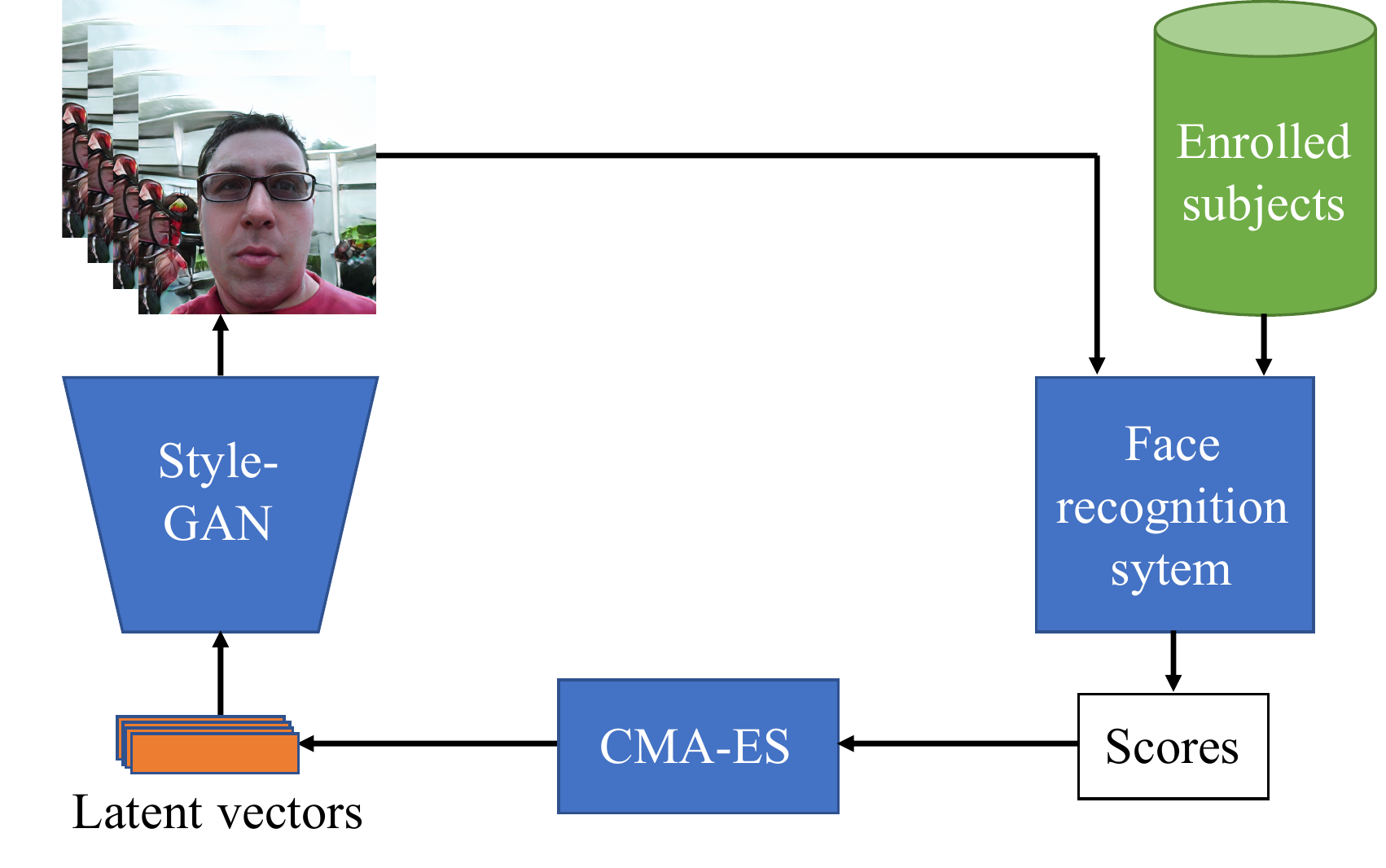}
	\caption{Overview of proposed method. With default setting, 22 latent vectors with a dimensionality of 512 are fed into StyleGAN~\cite{karras2019style} to generate 22 facial images. The surrogate FR system then calculates mean score for each image on basis of enrolled subjects. CMA-ES~\cite{hansen2001completely} algorithm then uses scores to estimate 22 new latent vectors.}
	\label{fig:lve}
\end{figure}

An overview of the proposed method is shown in Figure~\ref{fig:lve}. In addition to the LVE algorithm, we need a pre-trained face generator (in this case, StyleGAN~\cite{karras2019style}), a surrogate pre-trained FR system (used to approximate the target FR system), a surrogate face database, and an implementation of the CMA-ES algorithm~\cite{hansen2001completely}. If a pre-trained generator is not available, one must be trained from scratch:
\begin{itemize}
	\item Prepare three or four face databases: one to train StyleGAN, one or two to train the FR system (in the case of two, one is used for validation), and one to run the LVE algorithm. Some or all of them could overlap; in our work, we used the hardest case, i.e., non-overlapping, to demonstrate the generalizability of our proposed method.
	\item Prepare and train FR system.
	\item Prepare and train StyleGAN.
	\item Implement or use open-source library for CMA-ES algorithm, e.g., pycma~\cite{hansen2019pycma}.
	\item Run LVE algorithm.
\end{itemize}

\subsection{Latent Variable Evolution}

\begin{algorithm}
	\caption{Latent variable evolution.}
	\label{alg:lve}
	\begin{algorithmic}
		\State $m \gets 22$
		\Comment{Population size}
		
		\Procedure{RunLVE}{$m, n$}
		\State MasterFaces = \{\}
		\Comment{Master face set}
		\State MasterScores = \{\}
		\Comment{and the corresponding score set}
		
		\State $\textbf{z} \gets 0$
		\Comment{Initialize latent vectors $\bm{z} \in \R^{m}$}
		
		\For {$n$ iterations}
		\Comment{Run LVE algorithm $n$ times}
		\State $F \gets$ StyleGAN($\bm{z}$)
		\Comment{Generate $m$ faces $F$}
		
		\State $\bm{s} \gets 0$
		\Comment{Initialize scores $\mathbf{s} \in \R^{m}$}
		
		\For {face $F_i$ in faces $F$}
		\For{face $E_j$ in data $E$}
		\State $s_i \gets s_i +$ FaceMatching($F_i, E_j$)
		\EndFor
		
		\State $\bm{s} \gets \frac{\bm{s}}{|E|}$
		\Comment{Calculate the mean scores}
		
		\State $F_b, s_b \gets$ GetBestFace($F, \bm{s}$)
		
		\State MasterFaces.append($F_b$)
		%\Comment{Append best master face}
		\State MasterScores.append($s_b$)
		%\Comment{and its corresponding score}
		
		\State $\bm{z} \gets$ CMA\_ES($\bm{s}$)
		\Comment{Evolve $\bm{z}$ based on $\bm{s}$}
		\EndFor
		\EndFor
		
		\State \textbf{return} MasterFaces, MasterScores
		\EndProcedure
		
		\State $F_b, s_b \gets$ GetBestFace(MasterFaces, MasterScores)
		%\Comment{\textbf{Final selection}}
		
	\end{algorithmic}
\end{algorithm}

The LVE procedure is formalized in Algorithm~\ref{alg:lve}. The FaceMatching($F_i, E_j$) function calculate the similarity between two input faces $F_i$ and $E_j$. As the default setting for the CMA-ES library~\cite{hansen2019pycma}, we set population size $m$ to 22. Unlike the algorithm of Bontrager et al.~\cite{bontrager2018deepmasterprints}, our algorithm does not require information about the pre-defined false matching rate (FMR). Moreover, the accumulated matching scores $\bm{s}$ are not added in a binary way (matched or unmatched: FaceMatching($\cdot,\cdot$) function only returns 1 or 0, respectively) but instead by using the actual similarity scores calculated by the FR system. As a result, the CMA-ES algorithm tries to maximize the $\bm{s}$ in each iteration. Therefore, the optimization curve is smoother than that of the Bontrager's LVE algorithm, especially when the training data for StyleGAN, the FR system, and this LVE algorithm ($E$) differ. One example is that, if there are no matches, the accumulated score $s_i$ of each generated face $F_i$ is 0 with the Bontrager's algorithm, whereas the CMA-ES algorithm has no clue to use in evolving $\bm{z}$. This problem is solved by using the actual scores for $\bm{s}$.

The local best master face $F_b$ is selected among $m$ master faces and collected after each iteration on the basis of its score $s_b$, which is also logged. After $n$ iterations, the final (global) best master face is chosen among $n$ best master faces on the basis of the logged scores. The reason we do this instead of selecting the best master face of all master faces in every iteration is that (1) it reduces the storage of master faces and (2), if the number of iterations is large enough, besides getting better, the $m$ master faces generated in each iteration get closer to each other (by identity, appearance, background, and pose). Therefore, selecting the best one among them is sufficient. The running of the LVE algorithm on two databases based on the FMR and t-distributed stochastic neighbor embedding (t-SNE) visualization~\cite{maaten2008visualizing} of the master faces at certain iterations is described in section~\ref{sec:run_lve}.

\section{Evaluation}
\label{sec:eval}

In this section, we first describe our experimental design, including the FR systems and databases we used (section~\ref{sec:design}). We then describe the running of the proposed LVE algorithm on the LFW - Fold 1 database (scenario 1) and MOBIO database (scenario 2) and the analysis of its behavior (section~\ref{sec:run_lve}). Finally, we describe the calculation of the attack success probabilities of the obtained master faces for both scenario 1 (section \ref{sec:scen_1}) and 2 (section \ref{sec:scen_2}).

\subsection{Experimental Design}
\label{sec:design}
\subsubsection{Face Recognition Systems}

We used four pre-trained FR systems supported by the Bob toolbox~\cite{bob2017}:
\begin{itemize}
	\item Inception-ResNet-v2~\cite{szegedy2017inception} based FR systems trained by de Freitas Pereira et al.~\cite{de2018heterogeneous}: one trained on the CASIA-WebFace database~\cite{yi2014learning} and one trained on the MS-Celeb database~\cite{guo2016ms}. \textbf{We used the one trained on the CASIA-WebFace database to run the LVE algorithm}.
	
	\item FaceNet (using the Inception-v1 architecture~\cite{szegedy2015going}) proposed by Schroff et al.~\cite{schroff2015facenet}, implemented and trained by David Sandberg~\cite{sandberg2017facenet} on the MS-Celeb database~\cite{guo2016ms}.
	
	\item DR-GAN proposed and implemented by Tran et al.~\cite{tran2017disentangled}, pre-trained on a combination of the Multi-PIE database~\cite{gross2010multi} and the CASIA-WebFace database~\cite{yi2014learning}.
\end{itemize}

\subsubsection{Databases}

Beside the databases used to train the FR systems mentioned above (CASIA-WebFace~\cite{yi2014learning}, MS-Celeb~\cite{guo2016ms}, and Multi-PIE~\cite{gross2010multi}), we used the MOBIO database~\cite{mccool2012bi} with both male and female components and the LFW database~\cite{LFWTechUpdate} aligned by funneling~\cite{Huang2007a} to run the LVE algorithm and validate the master faces. The LFW database has several protocols; we used the fold 1 protocol. The MOBIO and LFW - Fold 1 databases were divided into two mutually exclusive sets (with non-overlapping identities): a world set used for training and a development (dev) set used for threshold selection for the FR systems (based on the equal error rate, EER), and an evaluation (eval) set. The dev and eval sets were both used to evaluate the master faces. In the dev and eval sets, the test pairs included both genuine and zero-effort imposter cases. To test the master faces, we replaced the test pairs by matching the master faces with all enrolled subjects and measured the false matching rates (FMRs).

The StyleGAN face generator was pre-trained on the Flickr-Faces-HQ (FFHQ) database~\cite{karras2019style}. For this training, there were no overlaps between the databases used for training the FR systems, training StyleGAN, and running the LVE algorithm. We wanted to demonstrate the generalizability of our proposed method since attackers often lack the knowledge and resources needed to perform this kind of attack and therefore tend to use resources widely available on the Internet.

\subsection{Running Latent Variable Evolution}
\label{sec:run_lve}

\begin{figure}[th!]
	\centering
	\includegraphics[width=80mm]{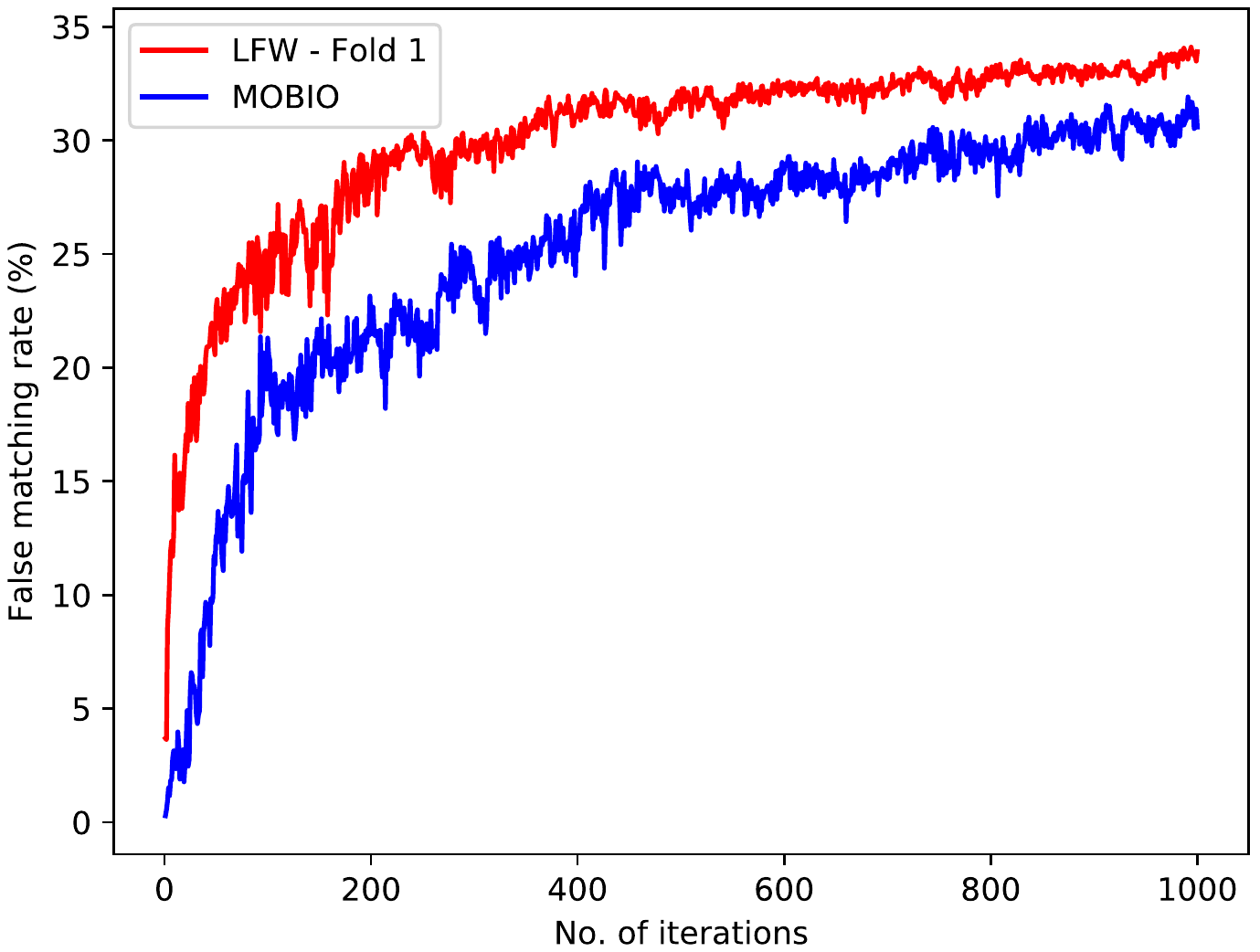}
	\caption{FMR for Inception-ResNet-v2 based FR system~\cite{szegedy2017inception} when running LVE algorithm on LFW - Fold 1 database~\cite{LFWTechUpdate} and MOBIO database~\cite{mccool2012bi}.}
	\label{fig:fmr}
\end{figure}

\begin{figure}[th!]
	\centering
	\includegraphics[width=82mm]{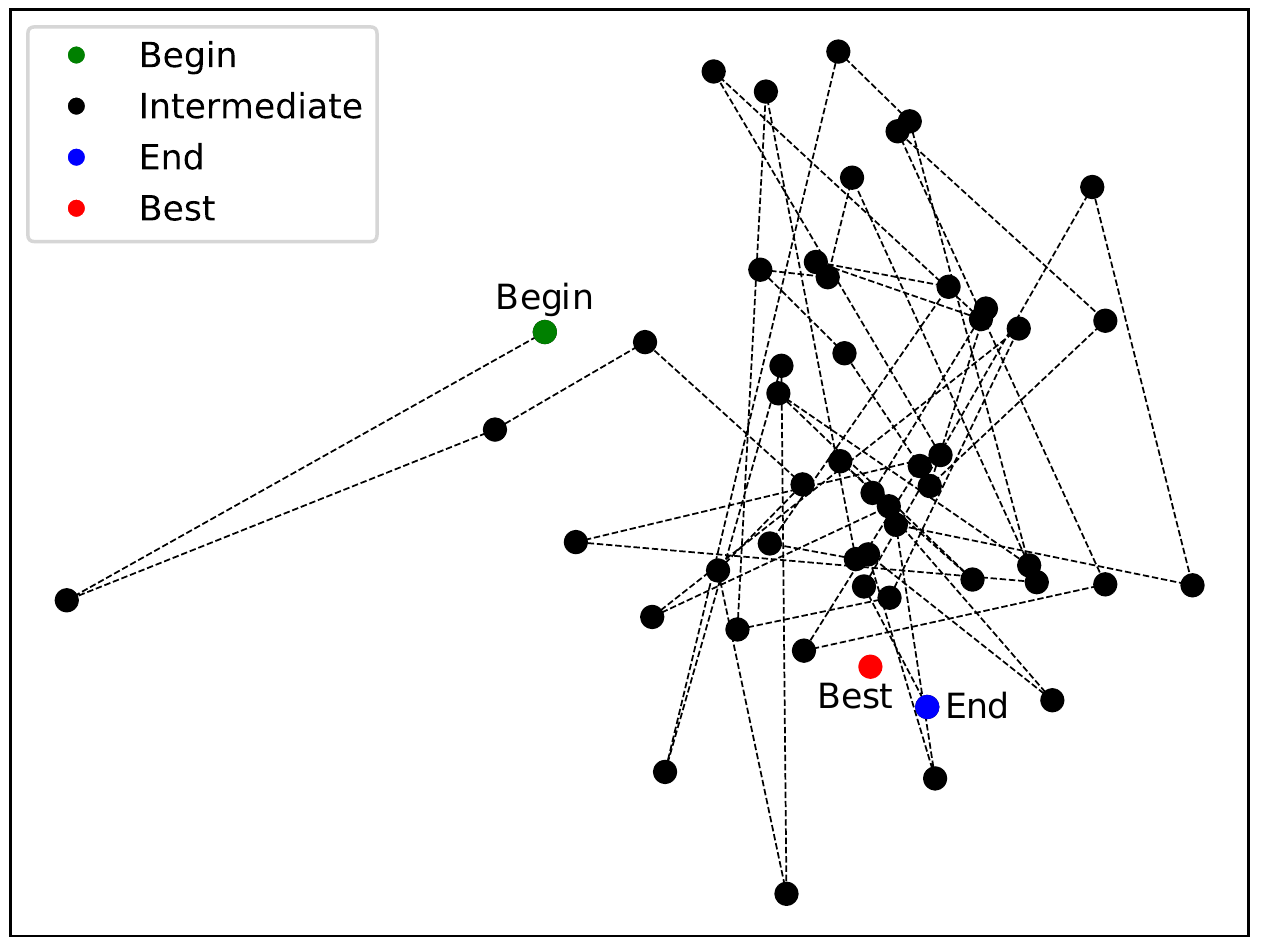}
	\caption{T-SNE visualization of master faces obtained every 20 iterations (1000 in total) on LFW - Fold 1 database~\cite{LFWTechUpdate}. Green dot represents master face at beginning; it is connected by dashed lines with intermediate master faces (black dots) that end at blue dot. Red dot represents best master face, created at 989~\textsuperscript{th} iteration; therefore, it does not overlap any black dot.}
	\label{fig:tsne}
\end{figure}

We ran the LVE algorithm on two databases (LFW - Fold 1~\cite{LFWTechUpdate} - scenario 1 and MOBIO~\cite{mccool2012bi} - scenario 2) with the Inception-ResNet-v2 based FR system~\cite{szegedy2017inception} trained on the CASIA-WebFace database~\cite{yi2014learning}. We ran it on a PC without a GPU for only 1000 iterations, which took less than 24 hours per database. The FMRs for the two databases are plotted in Figure~\ref{fig:fmr}. Since the MOBIO database has high variability in the pose and illumination conditions compared with the LFW - Fold 1 database, which greatly affected the FR system (the selected threshold was trickier), the FMRs for the MOBIO database were lower than those for the LFW database. They were about 35\% for the LFW - Fold 1 database and 30\% for the MOBIO database at the 1000\textsuperscript{th} iteration. Nevertheless, both curves still tend to increase. We limited the number of iterations to demonstrate that the attack can be done in a limited time. Comprehensive analysis will be done in follow-up work.

T-SNE visualization~\cite{maaten2008visualizing} of the process of running the LVE algorithm on the LFW - Fold 1 database is shown in Figure~\ref{fig:tsne}. Initially, the CMA-ES algorithm was unsure about the optimal direction. After finding some clues, it began generating master faces that jumped around the best master face (the red dot) and came closer and closer to it.

\subsection{Scenario 1: LFW - Fold 1 Database}
\label{sec:scen_1}

\begin{figure*}[th!]
	\centering
	\includegraphics[width=120mm]{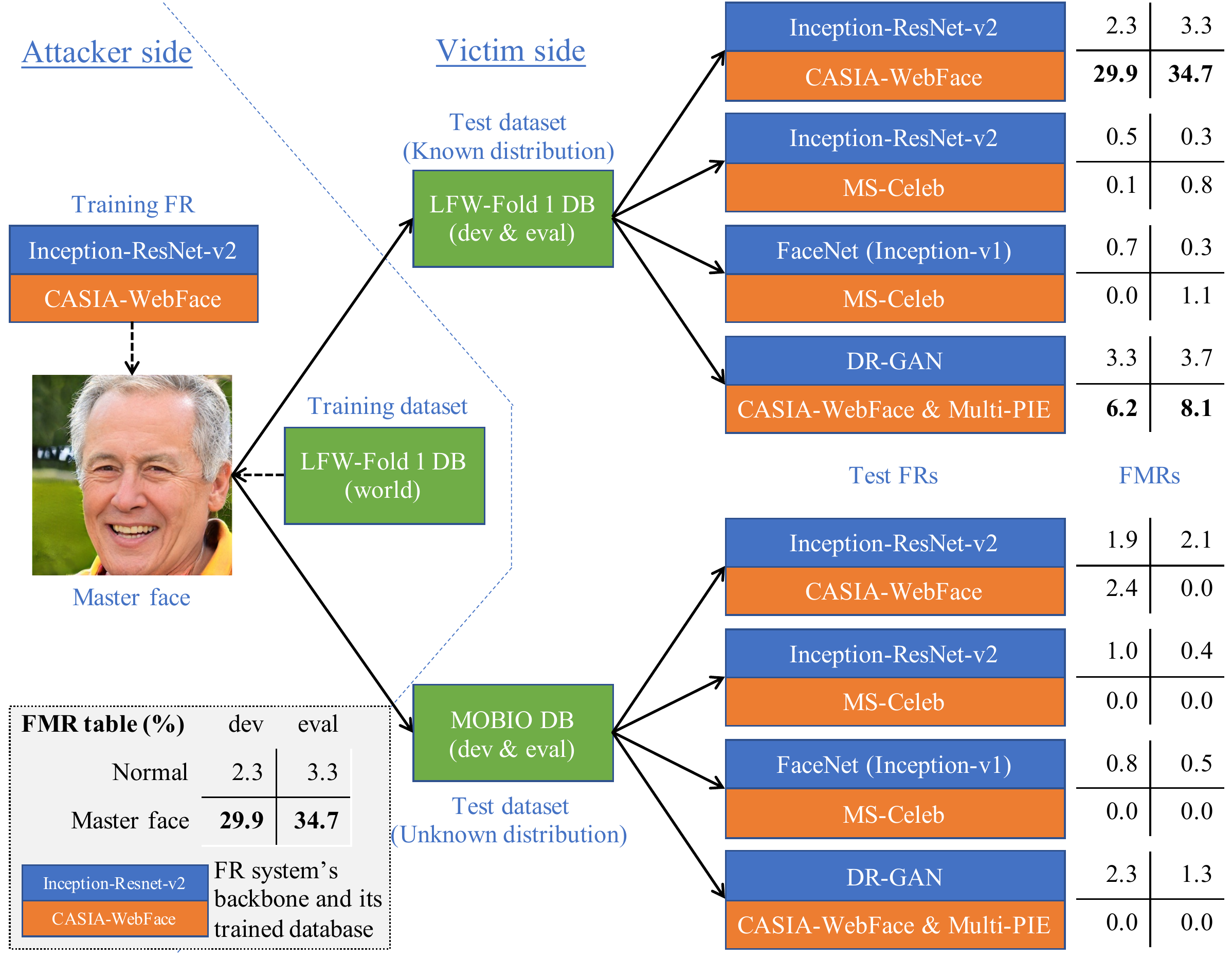}
	\caption{FMRs of original test designs and of master face generated using LFW - Fold 1 database~\cite{LFWTechUpdate} calculated using four configurations of three FR systems on LFW - Fold 1 database~\cite{LFWTechUpdate} and MOBIO database~\cite{mccool2012bi}.}
	\label{fig:lfw-fold1}
\end{figure*}

\begin{figure}[th!]
	\centering
	\includegraphics[width=83mm]{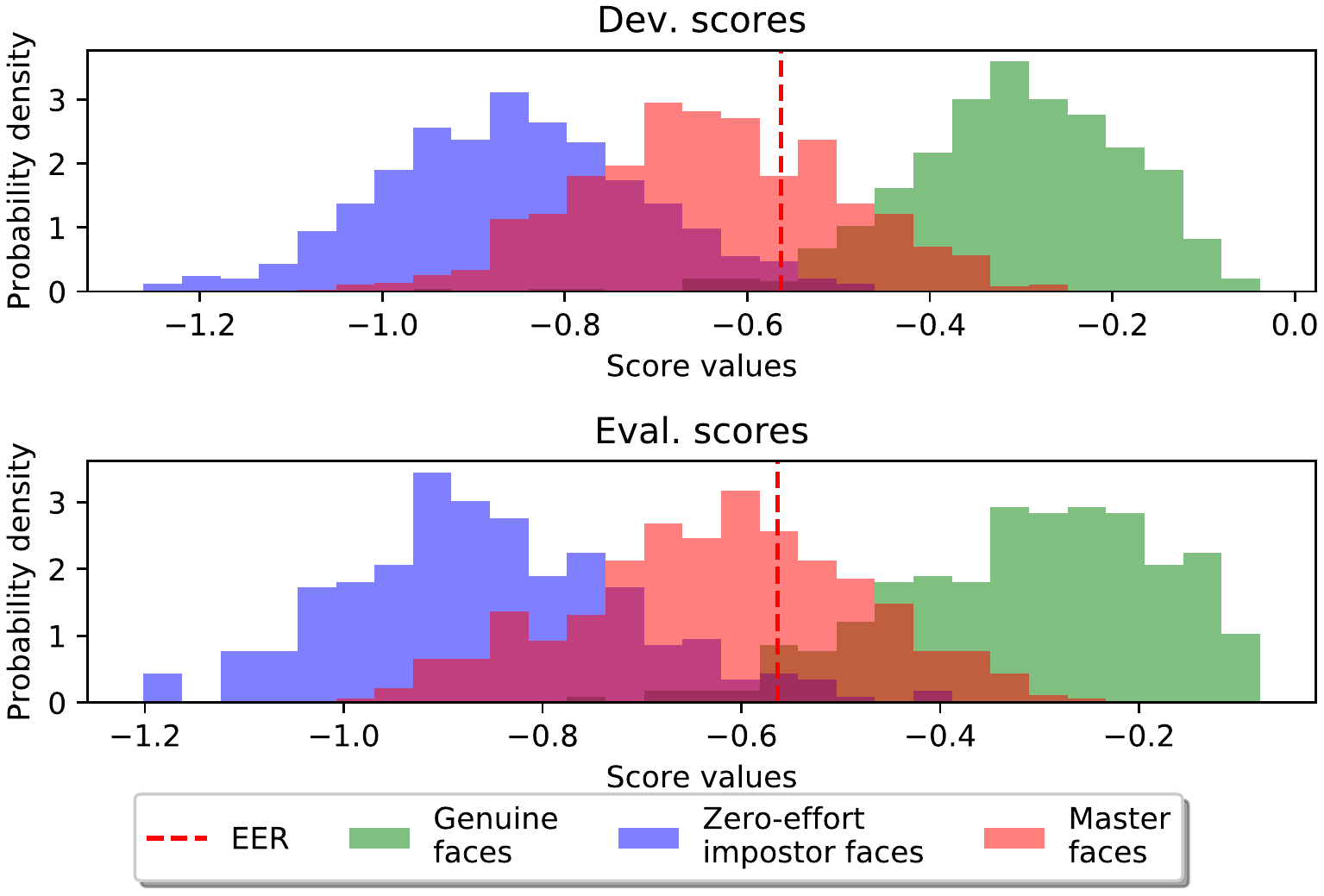}
	\caption{Histogram of scores for genuine faces, zero-effort imposter faces, and master face generated using LFW - Fold 1 database~\cite{LFWTechUpdate} calculated using Inception-ResNet-v2 based FR system~\cite{szegedy2017inception}.}
	\label{fig:hist}
\end{figure}

In this scenario, we ran the LVE algorithm on the LFW - Fold 1 database~\cite{LFWTechUpdate} using the Inception-ResNet-v2 based FR system~\cite{szegedy2017inception}. The best master face is shown in Figure~\ref{fig:masterfaces} (top right). We then tested it using three FR systems with four configurations on the LFW - Fold 1 database~\cite{LFWTechUpdate} and the MOBIO database~\cite{mccool2012bi}. The results are shown in Figure~\ref{fig:lfw-fold1} and summarized in Table~\ref{tab:summary}. 

We tried to match the obtained master face with all enrolled faces in the dev and eval sets of the LFW - Fold 1 database. A score histogram for the master face is plotted in Figure~\ref{fig:hist} along with those for the genuine faces and the zero-effort imposter faces from the original test design of the database. The master face scores moved away from the zero-effort imposter scores in the direction of the genuine face scores with about 30–35\% overlap. This means that the master face matched 30–35\% of the enrolled faces, which is significant.

As shown in Figure~\ref{fig:lfw-fold1}, the wolf attack worked on two FR systems on the LFW - Fold 1 database: the Inception-ResNet-v2 based one trained on the CASIA-WebFace database (which was also used to train the master face) and the DR-GAN one trained on a combination of the CASIA-WebFace and Multi-PIE databases. The results for the DR-GAN FR system demonstrate that the proposed method is generalizable to other FR system architectures. The wolf attack did not work in two cases:

\begin{itemize}
	\item LFW - Fold 1 database: The Inception-ResNet-v2 and Inception-v1 based FR systems were trained on the MS-Celeb database. Since the MS-Celeb database is larger than the CASIA-WebFace one, the FR systems trained on it were more robust than the others, which can be observed from the FMRs for the normal test cases without master faces. Since the master face was trained using the FR system trained on the CASIA-WebFace database, it was not strong enough to work on the normal cases.
	
	\item MOBIO database: As we mentioned above, the MOBIO database has high variability in the pose and illumination conditions compared with the LFW - Fold 1 database; therefore, the master face generated using the LFW - Fold 1 database was not sophisticated enough to work with the MOBIO database.
\end{itemize}

\subsection{Scenario 2: MOBIO Database}
\label{sec:scen_2}

\begin{figure*}[th!]
	\centering
	\includegraphics[width=120mm]{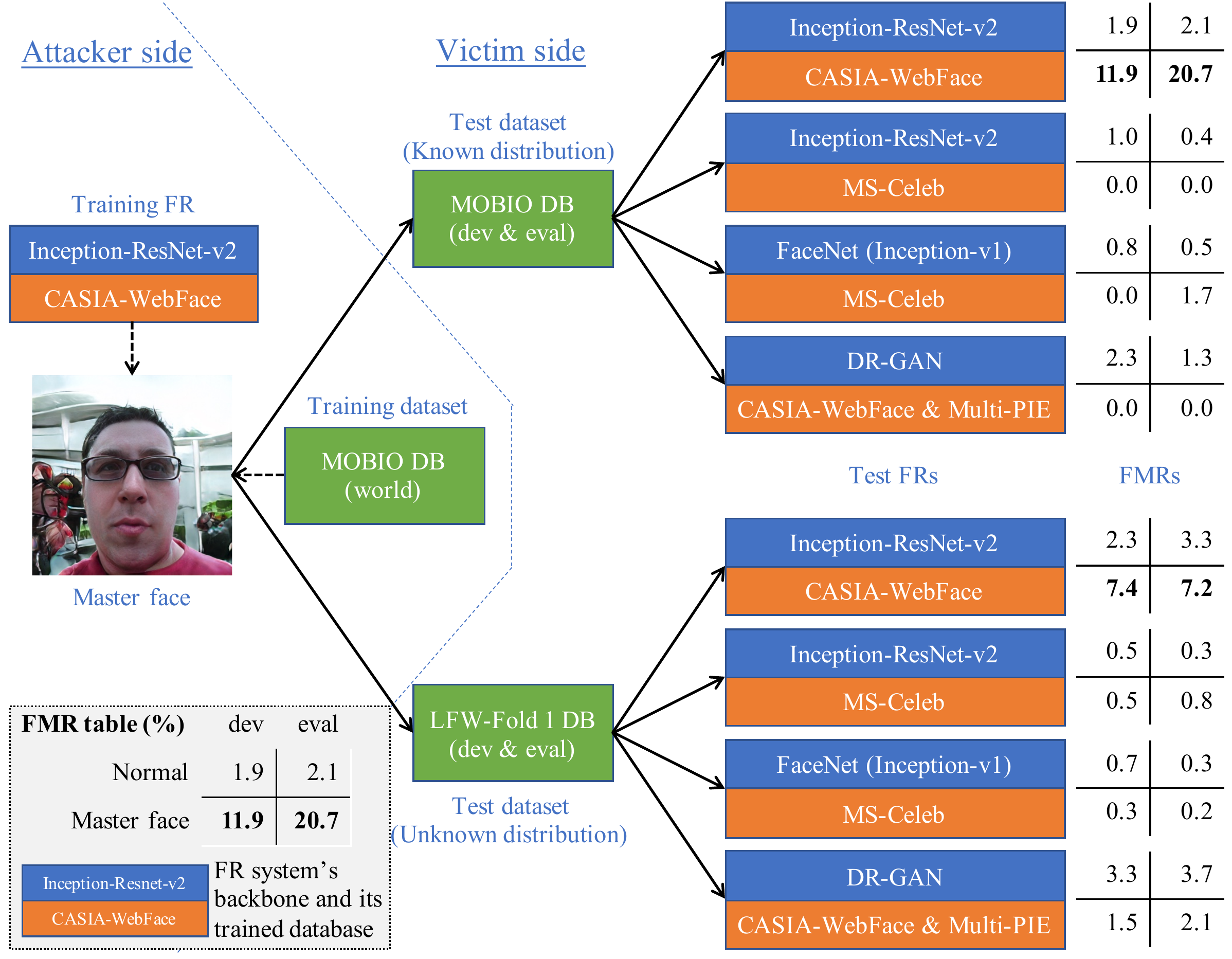}
	\caption{FMRs of original test designs and of master face generated using MOBIO database~\cite{mccool2012bi} calculated using four configurations of three FR systems on MOBIO database~\cite{mccool2012bi} and LFW - Fold 1 database~\cite{LFWTechUpdate}.}
	\label{fig:mobio}
\end{figure*}

\begin{table*}[th!]
	\centering
	\adjustbox{max width=\textwidth}{
		\begin{tabular}{l|c|c|cc}
			\multicolumn{1}{c|}{\multirow{2}{*}{\textbf{FR Setting}}} & \multicolumn{2}{c|}{\textbf{Scenario 1}} & \multicolumn{2}{c}{\textbf{Scenario 2}} \\ \cline{2-5} 
			\multicolumn{1}{c|}{} & \textbf{Known DB} & \textbf{Unknown DB} & \multicolumn{1}{c|}{\textbf{Known DB}} & \textbf{Unknown DB} \\ \hline
			Same Arch. - Same DB & \textbf{1} & 0 & \multicolumn{1}{c|}{\textbf{1}} & \textbf{1} \\
			Same Arch. - Different DB & 0 & 0 & \multicolumn{1}{c|}{0} & 0 \\
			Different Arch. - Same DB & \textbf{1} & 0 & \multicolumn{1}{c|}{0} & 0 \\
			Different Arch. - Different DB & 0 & 0 & \multicolumn{1}{c|}{0} & 0
	\end{tabular}}
	\caption{\label{tab:summary}Summary of successful attacks for scenarios 1 and 2 with different FR system settings (including their architecture and database used to train them) and databases.}
\end{table*}

In this scenario, we ran the LVE algorithm on the MOBIO database~\cite{mccool2012bi} using the Inception-ResNet-v2 based FR system~\cite{szegedy2017inception}. The best master face is shown in Figure~\ref{fig:masterfaces} (top left). As in the previous scenario, we tested it using three FR systems with four configuration on the MOBIO database~\cite{mccool2012bi} and the LFW - Fold 1 database~\cite{LFWTechUpdate}. The results are shown in Figure~\ref{fig:mobio} and are summarized in Table~\ref{tab:summary}. 

The wolf attack worked on the Inception-ResNet-v2 based FR system trained on the CASIA-WebFace database (which was also used to train the master face) on both the MOBIO and LFW - Fold 1 databases. The FMRs were about 12–20\% on the MOBIO database and 7\%–12\% on the LFW - Fold 1 database. These results demonstrate that the proposed method is generalizable to different databases. This is because the MOBIO database is more sophisticated than the LFW - Fold 1 database. Unfortunately, it did not work on the other FR systems. One possible explanation is that since the MOBIO database is sophisticated, the LVE algorithm was trying to ``overfit'' the used FR system with master faces that were uncommon to other FR systems to increase the FMRs.

\begin{figure}[th!]
	\centering
	\includegraphics[width=\columnwidth]{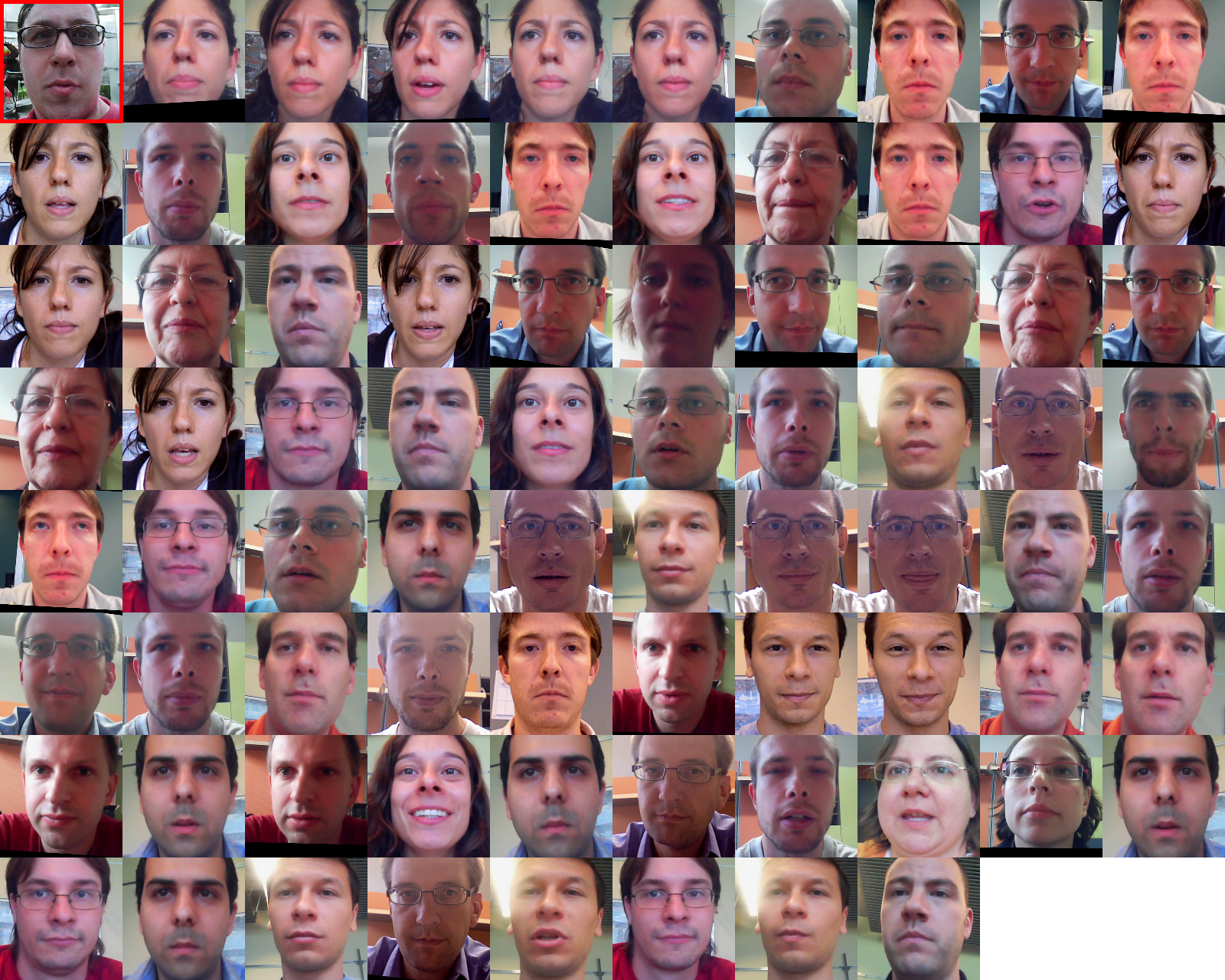}
	\caption{Master face (bordered in red) and all matching enrolled faces sorted from nearest to furthest for eval set of MOBIO database~\cite{mccool2012bi}.}
	\label{fig:matches_mobio}
\end{figure}

\begin{figure}[th!]
	\centering
	\includegraphics[width=\columnwidth]{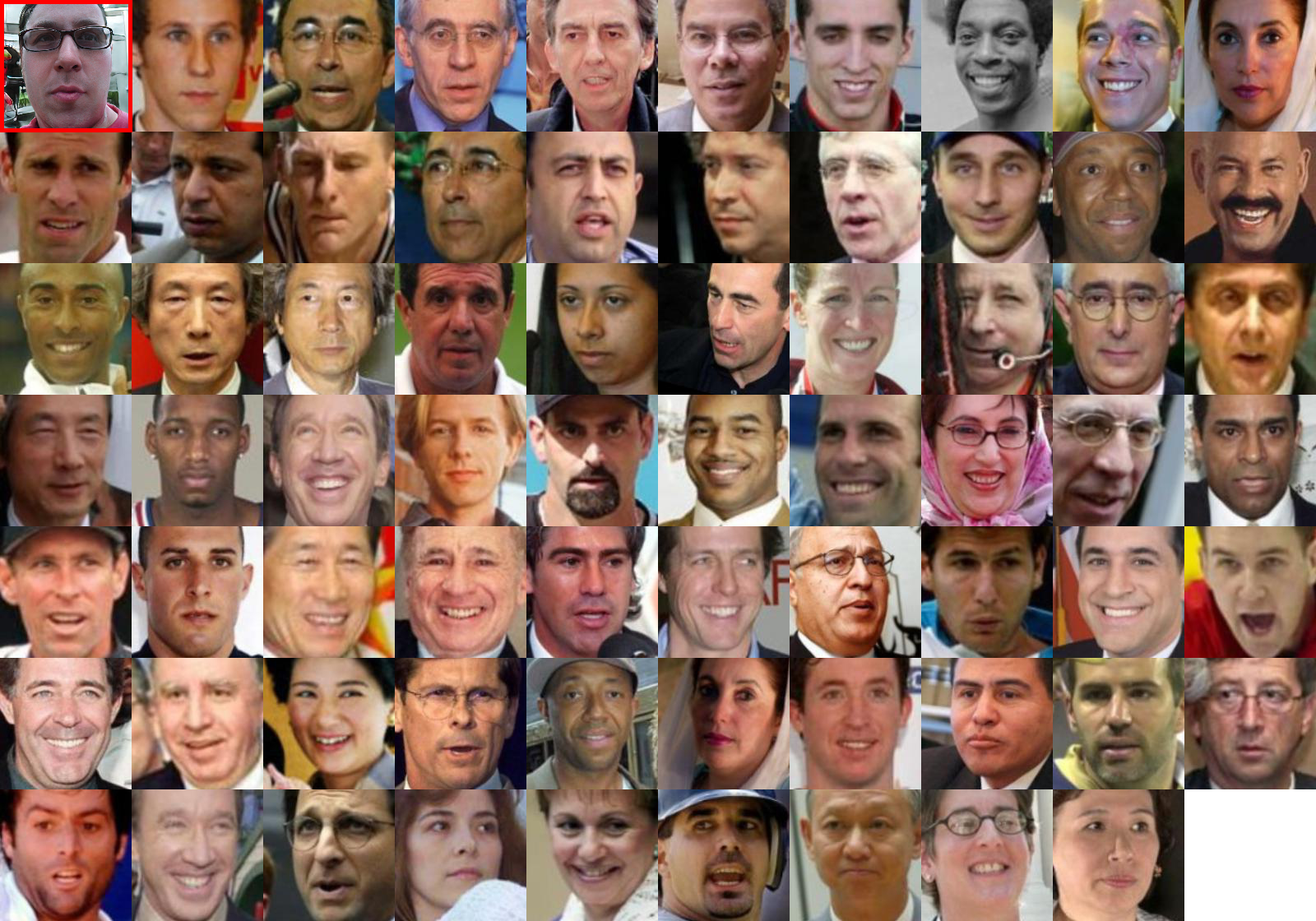}
	\caption{Master face (bordered in red) and all matching enrolled faces sorted from nearest to furthest for dev set of LFW - Fold 1 database~\cite{LFWTechUpdate}.}
	\label{fig:matches_lfw-fold1}
\end{figure}

Two examples of all faces matched are shown in Figure~\ref{fig:matches_mobio} (eval set for MOBIO database) and Figure~\ref{fig:matches_lfw-fold1} (dev set for LFW - Fold 1 database). The images were sorted from nearest to furthest. The master face matched both male and female subjects of different races with different skin tones, poses, and illumination, with or without beard, glasses, hat, and scarf. The master face was trained on the MOBIO database, which has many selfie-like photos; therefore, it also had an appearance of a selfie photo.

In addition to helping us understand the attacker's point of view, the obtained results can be used to identify the weaknesses of an FR system. For example, the Inception-ResNet-v2 based FR system trained on the CASIA-WebFace database has two weaknesses: it has trouble distinguishing between male and female and understanding racial differences. Another example comes from Fig~\ref{fig:masterfaces} shown in section~\ref{sec:intro}. The Inception-ResNet-v2 based FR system trained on the MS-Celeb database seems to be poor at recognizing images of children as two of the master faces are of children. One possible explanation is that the MS-Celeb database lacks photographs of children since most celebrities are teenagers or adults.

\section{Defense Against Master Face Attack}
\label{sec:defense}

To prevent such kinds of attack, besides improving FR systems, we need to use an additional detector to filter out master faces. For camera-based FR and face authentication systems, using a presentation attack detector~\cite{bhattacharjee2019recent} is a good option. If the system takes a digital image as input, a computer-generated/manipulated image detector~\cite{verdoliva2020media} is needed. However, the generalizability of such detectors is a major concern~\cite{khodabakhsh2018fake}, especially when the master faces were generated using databases that have a different distribution from those covered by the detectors. Although recent work has addressed this problem, performance on cross-databases is still limited~\cite{nguyen2019multi,mohammadi2020domain,mohammadi2020improving} and needs further improvement.

\section{Summary and Future Work}
\label{sec:sum}
Aimed at simplicity by using available resources easily obtained on the Internet, including a pre-trained StyleGAN model, a pre-trained face recognition system, a face database, and a conventional PC without a GPU, our proposed method can generate master faces in less than a day. Beside the ability to attack seen data and seen face recognition systems (white box attacks), the master faces, in some cases, can be generalized. This discovery raises concerns about the robustness of face recognition and face authentication systems. Moreover, the properties of the master faces can provide clues for understanding and improving FR systems. Although countermeasures can be used to mitigate this type of attack, further research is needed to make face authentication systems more secure, especially when they are used in applications related to finance and personal data.

% Future work will mainly focus on deep analysis of the properties of the master faces on several face recognition systems and databases, on using multiple surrogate FR systems for LVE, and on improving the generalizability of the proposed method on cross-platform FR systems and cross-databases. The recently released StyleGAN 2~\cite{karras2019analyzing} will be used instead of the previous version to improve the quality of the generated images.

Future work will mainly focus on deep analysis of the properties of the master faces, mostly about the correlation between the skin color, race, gender, age, and pose of the master face and their proportion in the data used for running LVE. Another important task is to perform more experiments on more face recognition systems and more databases and to improve the generalizability of the master faces on them. One possible solution for the generalizability is using multiple surrogate face recognition systems for LVE. The recently released StyleGAN 2~\cite{karras2019analyzing} will be used instead of the previous version to improve the quality of the generated images.

\section*{Acknowledgements}

This research was supported by JSPS KAKENHI Grants JP16H06302, JP17H04687, JP18H04120, JP18H04112, JP18KT0051, and by JST CREST Grant JPMJCR18A6, Japan.

We would like to thank Dr. Tiago de Freitas Pereira and Dr. Amir Mohammadi from Biometrics Security and Privacy (BSP) group at Idiap for providing the pre-trained face recognition systems and for their supports on Bob toolkit.

{\small
	\bibliographystyle{ieee}
	\bibliography{refs}

\begin{thebibliography}{10}\itemsep=-1pt

\bibitem{bob2017}
A.~Anjos, M.~G\"unther, T.~de~Freitas~Pereira, P.~Korshunov, A.~Mohammadi, and
  S.~Marcel.
\newblock Continuously reproducing toolchains in pattern recognition and
  machine learning experiments.
\newblock In {\em ICML}, Aug. 2017.

\bibitem{arjovsky2017wasserstein}
M.~Arjovsky, S.~Chintala, and L.~Bottou.
\newblock Wasserstein generative adversarial networks.
\newblock In {\em ICML}, pages 214--223, 2017.

\bibitem{bhattacharjee2019recent}
S.~Bhattacharjee, A.~Mohammadi, A.~Anjos, and S.~Marcel.
\newblock Recent advances in face presentation attack detection.
\newblock In {\em Handbook of Biometric Anti-Spoofing}, pages 207--228.
  Springer, 2019.

\bibitem{bontrager2018deep}
P.~Bontrager, W.~Lin, J.~Togelius, and S.~Risi.
\newblock Deep interactive evolution.
\newblock In {\em International Conference on Computational Intelligence in
  Music, Sound, Art and Design}, pages 267--282. Springer, 2018.

\bibitem{bontrager2018deepmasterprints}
P.~Bontrager, A.~Roy, J.~Togelius, N.~Memon, and A.~Ross.
\newblock Deepmasterprints: Generating masterprints for dictionary attacks via
  latent variable evolution.
\newblock In {\em BTAS}, pages 1--9. IEEE, 2018.

\bibitem{de2018heterogeneous}
T.~de~Freitas~Pereira, A.~Anjos, and S.~Marcel.
\newblock Heterogeneous face recognition using domain specific units.
\newblock {\em IEEE Transactions on Information Forensics and Security},
  14(7):1803--1816, 2018.

\bibitem{goodfellow2014generative}
I.~Goodfellow, J.~Pouget-Abadie, M.~Mirza, B.~Xu, D.~Warde-Farley, S.~Ozair,
  A.~Courville, and Y.~Bengio.
\newblock Generative adversarial nets.
\newblock In {\em NIPS}, pages 2672--2680, 2014.

\bibitem{gross2010multi}
R.~Gross, I.~Matthews, J.~Cohn, T.~Kanade, and S.~Baker.
\newblock Multi-{PIE}.
\newblock {\em Image and Vision Computing}, 28(5):807--813, 2010.

\bibitem{gulrajani2017improved}
I.~Gulrajani, F.~Ahmed, M.~Arjovsky, V.~Dumoulin, and A.~C. Courville.
\newblock Improved training of wasserstein {GANs}.
\newblock In {\em NIPS}, pages 5767--5777, 2017.

\bibitem{guo2016ms}
Y.~Guo, L.~Zhang, Y.~Hu, X.~He, and J.~Gao.
\newblock Ms-celeb-1m: A dataset and benchmark for large-scale face
  recognition.
\newblock In {\em ECCV}, pages 87--102. Springer, 2016.

\bibitem{hansen2019pycma}
N.~Hansen, Y.~Akimoto, and P.~Baudis.
\newblock {CMA-ES/pycma} on {G}ithub.
\newblock Zenodo, DOI:10.5281/zenodo.2559634, Feb. 2019.

\bibitem{hansen2001completely}
N.~Hansen and A.~Ostermeier.
\newblock Completely derandomized self-adaptation in evolution strategies.
\newblock {\em Evolutionary computation}, 9(2):159--195, 2001.

\bibitem{Huang2007a}
G.~B. Huang, V.~Jain, and E.~Learned-Miller.
\newblock Unsupervised joint alignment of complex images.
\newblock In {\em ICCV}. IEEE, 2007.

\bibitem{karras2018progressive}
T.~Karras, T.~Aila, S.~Laine, and J.~Lehtinen.
\newblock Progressive growing of {GANs} for improved quality, stability, and
  variation.
\newblock In {\em ICLR}, 2018.

\bibitem{karras2019style}
T.~Karras, S.~Laine, and T.~Aila.
\newblock A style-based generator architecture for generative adversarial
  networks.
\newblock In {\em CVPR}, pages 4401--4410. IEEE, 2019.

\bibitem{karras2019analyzing}
T.~Karras, S.~Laine, M.~Aittala, J.~Hellsten, J.~Lehtinen, and T.~Aila.
\newblock Analyzing and improving the image quality of stylegan.
\newblock {\em arXiv preprint arXiv:1912.04958}, 2019.

\bibitem{khodabakhsh2018fake}
A.~Khodabakhsh, R.~Ramachandra, K.~Raja, P.~Wasnik, and C.~Busch.
\newblock Fake face detection methods: Can they be generalized?
\newblock In {\em BIOSIG}, pages 1--6. IEEE, 2018.

\bibitem{kingma2014auto}
D.~P. Kingma and M.~Welling.
\newblock Auto-encoding variational bayes.
\newblock In {\em ICLR}, 2014.

\bibitem{LFWTechUpdate}
G.~B. H.~E. Learned-Miller.
\newblock Labeled faces in the wild: Updates and new reporting procedures.
\newblock Technical Report UM-CS-2014-003, University of Massachusetts,
  Amherst, May 2014.

\bibitem{maaten2008visualizing}
L.~v.~d. Maaten and G.~Hinton.
\newblock Visualizing data using t-sne.
\newblock {\em Journal of machine learning research}, 9(Nov):2579--2605, 2008.

\bibitem{mccool2012bi}
C.~McCool, S.~Marcel, A.~Hadid, M.~Pietik{\"a}inen, P.~Matejka,
  J.~Cernock{\`y}, N.~Poh, J.~Kittler, A.~Larcher, C.~Levy, et~al.
\newblock Bi-modal person recognition on a mobile phone: using mobile phone
  data.
\newblock In {\em ICMEW}, pages 635--640. IEEE, 2012.

\bibitem{mohammadi2020domain}
A.~Mohammadi, S.~Bhattacharjee, and S.~Marcel.
\newblock Domain adaptation for generalization of face presentation attack
  detection in mobile settings with minimal information.
\newblock In {\em ICASSP}. IEEE, 2020.

\bibitem{mohammadi2020improving}
A.~Mohammadi, S.~Bhattacharjee, and S.~Marcel.
\newblock Improving cross-dataset performance of face presentation attack
  detection systems using face recognition datasets.
\newblock In {\em ICASSP}. IEEE, 2020.

\bibitem{nguyen2019multi}
H.~H. Nguyen, F.~Fang, J.~Yamagishi, and I.~Echizen.
\newblock Multi-task learning for detecting and segmenting manipulated facial
  images and videos.
\newblock In {\em BTAS}. IEEE, 2019.

\bibitem{parkhi2015deep}
O.~M. Parkhi, A.~Vedaldi, and A.~Zisserman.
\newblock Deep face recognition.
\newblock In {\em BMVC}, pages 41.1--41.12. British Machine Vision Association,
  2015.

\bibitem{ratha2001enhancing}
N.~K. Ratha, J.~H. Connell, and R.~M. Bolle.
\newblock Enhancing security and privacy in biometrics-based authentication
  systems.
\newblock {\em IBM systems Journal}, 40(3):614--634, 2001.

\bibitem{roy2019masterprint}
A.~Roy, N.~Memon, and A.~Ross.
\newblock Masterprint attack resistance: A maximum cover based approach for
  automatic fingerprint template selection.
\newblock In {\em BTAS}. IEEE, 2019.

\bibitem{ILSVRC15}
O.~Russakovsky, J.~Deng, H.~Su, J.~Krause, S.~Satheesh, S.~Ma, Z.~Huang,
  A.~Karpathy, A.~Khosla, M.~Bernstein, A.~C. Berg, and L.~Fei-Fei.
\newblock {ImageNet Large Scale Visual Recognition Challenge}.
\newblock {\em International Journal of Computer Vision}, 2015.

\bibitem{sandberg2017facenet}
D.~Sandberg.
\newblock Facenet: face recognition using tensorflow.
\newblock https://github.com/davidsandberg/facenet, 2017.

\bibitem{scherhag2019face}
U.~Scherhag, C.~Rathgeb, J.~Merkle, R.~Breithaupt, and C.~Busch.
\newblock Face recognition systems under morphing attacks: A survey.
\newblock {\em IEEE Access}, 7:23012--23026, 2019.

\bibitem{schroff2015facenet}
F.~Schroff, D.~Kalenichenko, and J.~Philbin.
\newblock Facenet: A unified embedding for face recognition and clustering.
\newblock In {\em CVPR}, pages 815--823. IEEE, 2015.

\bibitem{simonyan2014very}
K.~Simonyan and A.~Zisserman.
\newblock Very deep convolutional networks for large-scale image recognition.
\newblock {\em arXiv preprint arXiv:1409.1556}, 2014.

\bibitem{szegedy2017inception}
C.~Szegedy, S.~Ioffe, V.~Vanhoucke, and A.~A. Alemi.
\newblock Inception-v4, inception-resnet and the impact of residual connections
  on learning.
\newblock In {\em AAAI}, 2017.

\bibitem{szegedy2015going}
C.~Szegedy, W.~Liu, Y.~Jia, P.~Sermanet, S.~Reed, D.~Anguelov, D.~Erhan,
  V.~Vanhoucke, and A.~Rabinovich.
\newblock Going deeper with convolutions.
\newblock In {\em CVPR}, pages 1--9, 2015.

\bibitem{tran2017disentangled}
L.~Tran, X.~Yin, and X.~Liu.
\newblock Disentangled representation learning {GAN} for pose-invariant face
  recognition.
\newblock In {\em CVPR}, pages 1415--1424. IEEE, 2017.

\bibitem{une2007wolf}
M.~Une, A.~Otsuka, and H.~Imai.
\newblock Wolf attack probability: A new security measure in biometric
  authentication systems.
\newblock In {\em ICB}, pages 396--406. Springer, 2007.

\bibitem{verdoliva2020media}
L.~Verdoliva.
\newblock Media forensics and deepfakes: an overview.
\newblock {\em arXiv preprint arXiv:2001.06564}, 2020.

\bibitem{wolberg1998image}
G.~Wolberg.
\newblock Image morphing: a survey.
\newblock {\em The visual computer}, 14(8-9):360--372, 1998.

\bibitem{wu2018light}
X.~Wu, R.~He, Z.~Sun, and T.~Tan.
\newblock A light {CNN} for deep face representation with noisy labels.
\newblock {\em IEEE Transactions on Information Forensics and Security},
  13(11):2884--2896, 2018.

\bibitem{yi2014learning}
D.~Yi, Z.~Lei, S.~Liao, and S.~Z. Li.
\newblock Learning face representation from scratch.
\newblock {\em arXiv preprint arXiv:1411.7923}, 2014.

\end{thebibliography}
}

\end{document}